%% file: main.tex
\PassOptionsToPackage{table,xcdraw}{xcolor}
\documentclass[sigconf]{acmart}
\usepackage{tikz}
\usepackage{svg}
\usepackage{algorithm}
\usepackage{algpseudocode}
\usepackage{booktabs, multirow}
\usepackage{soul}

\AtBeginDocument{
  }

\setcopyright{acmcopyright}
\copyrightyear{2022}
\acmYear{2022}
\acmDOI{XXXXXXX.XXXXXXX}

\newcommand{\true}{\ensuremath{\mathit{true}}}
\newcommand{\false}{\ensuremath{\mathit{false}}}

\newcommand{\myAnd}{\ensuremath{\wedge}}

\newcommand{\XOR}{\ensuremath{\mathit{XOR}}}
\newcommand{\myIff}{\ensuremath{\leftrightarrow}}

\begin{document}

\title{BNSynth: Bounded Boolean Functional Synthesis}

\author{Ravi Raja}
\email{raviraja@iisc.ac.in}
\affiliation{%
  \institution{Indian Institute of Science}
  \city{Bengaluru}
  \country{India}
}

\author{Stanly Samuel}
\email{stanly@iisc.ac.in}
\affiliation{%
  \institution{Indian Institute of Science}
  \city{Bengaluru}
  \country{India}
}

\author{Chiranjib Bhattacharyya}
\email{chiru@iisc.ac.in}
\affiliation{%
  \institution{Indian Institute of Science}
  \city{Bengaluru}
  \country{India}
}
 
\author{Deepak D'Souza}
\email{deepakd@iisc.ac.in}
\affiliation{%
  \institution{Indian Institute of Science}
  \city{Bengaluru}
  \country{India}
}

\author{Aditya Kanade}
\email{kanadeaditya@microsoft.com}
\affiliation{%
  \institution{Microsoft Research}
  \city{Bengaluru}
  \country{India}
}

\renewcommand{\shortauthors}{Raja et al.}

\begin{abstract}
The automated synthesis of correct-by-construction Boolean functions from logical specifications is known as the Boolean Functional Synthesis (BFS) problem. BFS has many application areas that range from software engineering to circuit design. In this paper, we introduce a tool BNSynth, that is the first to solve the BFS problem under a given bound on the solution space. Bounding the solution space induces the synthesis of smaller functions that benefit resource constrained areas such as circuit design. BNSynth uses a counter-example guided, neural approach to solve the bounded BFS problem. Initial results show promise in synthesizing smaller solutions; we observe at least \textbf{3.2X} (and up to \textbf{24X}) improvement in the reduction of solution size on average, as compared to state of the art tools on our benchmarks. BNSynth is available on GitHub under an open source license. \\ 
\textbf{Tool Link:} \url{https://github.com/rrlcs/bnsynth} \\
\textbf{Video Link:} \url{https://youtu.be/xaaopov3eZc}
\end{abstract}




\maketitle

\input{sections/intro.tex}

\section{Background}
\input{sections/bfs.tex}

\input{sections/gcln.tex}

\input{sections/tool-architecture}
\input{sections/evaluation}

\section{Future Work}
Although we show promise on the custom and LUT benchmarks, they do not reflect BFS applications that range over a large input domain. Scalability still remains a major challenge and is a work in progress.

\section{Conclusion}
In this paper, we introduced the Bounded BFS problem and made
preliminary progress towards solving the problem. We also introduce a
novel counterexample guided, neural based approach that significantly
outperforms a state of the art tool in formula size for small
specification sizes. This empirically shows that an efficient and
scalable solution to the B-BFS problem can provide significant savings
to resource constrained BFS domains.

\bibliographystyle{ACM-Reference-Format}
\bibliography{references}

\appendix

\end{document}

%% file: sections/intro.tex
\section{Introduction}

Boolean Functional Synthesis (BFS) is a well-known problem in the
domain of automated program synthesis from logical
specifications.
In this problem we are given a Boolean function $F(X,Y)$ over
sets of variables $X$ and $Y$, and we are asked to synthesize a Boolean
expression $\psi_i(X)$ for each $y_i \in Y$, such that, for any valuation
$v$ to $X$, if there exists a valuation to $Y$ which makes $F$ true, 
then the expression $F[\psi_i(X)/y_i]$ (where each $y_i$ is replaced
by $\psi_i(X)$) also evaluates to true under $v$.
%% This problem aims to synthesize a Boolean function
%% that is correct-by-construction with respect to the declared
%% specification; this specification symbolically relates the inputs and
%% outputs of the function to be synthesized.
Since Boolean functions are
the basic building blocks of modern digital systems, BFS has
applications in a wide range of areas, including QBF-SAT solving,
circuit repair and debugging. This has motivated the community to
develop practically efficient algorithms for synthesizing % compact
solutions to the BFS problem % Boolean functions
\cite{Manthan,akshay}. % , which is a non-trivial endeavor.
In many applications (like circuit repair) however, the \emph{compactness} of
the synthesized expressions is an important aspect.
To the best of our knowledge, current
techniques are unable to specify a bound on the Boolean function size
during synthesis.
% Specifying a bound on the size of the formula offers
% flexibility in synthesizing minimal-sized Boolean functions.
Our aim in this paper is to address the problem of finding compact
solutions to the BFS problem, using a neural synthesis approach.

Learning Boolean functions from logical specifications using neural
networks is a difficult problem as it requires the network to
represent Boolean functions. Boolean functions are discrete functions
and consequently, non-differentiable. Thus, learning a Boolean
function directly using traditional neural networks is not
possible. Recently Ryan et al proposed the Gated Continuous Logic
Network (GCLN) model \cite{gcln} that builds on Fuzzy Logic to
represent Boolean and linear integer operator, in the context of
learning invariants for programs. In this work, we investigate the use
of the GCLN model to synthesize solutions to the BFS problem. Our
model lets us bound the number of clauses used in the synthesized
Boolean expression.

We implement this approach in our tool BNSynth (for Bounded Neural
Synthesis), that also uses sampling and counterexample guided
techniques to synthesize Boolean functions. We validate our hypothesis
that this system can learn smaller expressions as compared to a
state-of-the-art tool, over custom and standard benchmarks. We observe
a 24.2X (resp.\@ 3.2) average improvement in the number of clauses and
35.1X (resp.\@ 4.2) improvement in literal count, for the custom
(resp.\@ standard) benchmarks. We also observe that in some cases, we
use fewer input lines. This empirically shows that our system is
capable of synthesizing smaller and efficient Boolean expressions as
compared to the state-of-the-art.

%% file: sections/bfs.tex
\subsection{Boolean Functional Synthesis (BFS)}
\label{sec:bfs}

\begin{figure*}[htp!]
  \begin{center}
    \begin{picture}(0,0)%
      \includegraphics{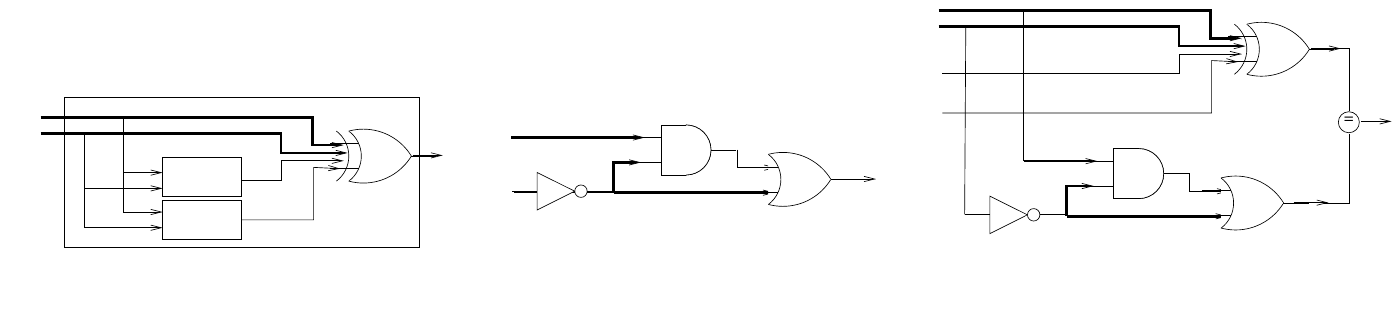}%
      \end{picture}%
      \setlength{\unitlength}{1657sp}%
      \begingroup\makeatletter\ifx\SetFigFont\undefined%
      \gdef\SetFigFont#1#2#3#4#5{%
        \reset@font\fontsize{#1}{#2pt}%
        \fontfamily{#3}\fontseries{#4}\fontshape{#5}%
        \selectfont}%
      \fi\endgroup%
      \begin{picture}(15918,3540)(4441,-4360)
      \put(11658,-4273){\makebox(0,0)[b]{\smash{{\SetFigFont{5}{6.0}{\rmdefault}{\mddefault}{\updefault}{\color[rgb]{0,0,0}(b) Specification $H(x_1,x_2)$}%
      }}}}
      \put(17260,-4291){\makebox(0,0)[b]{\smash{{\SetFigFont{5}{6.0}{\rmdefault}{\mddefault}{\updefault}{\color[rgb]{0,0,0}(c) BFS instance $F(x_1,x_2,y_1,y_2)$}%
      }}}}
      \put(6751,-2851){\makebox(0,0)[b]{\smash{{\SetFigFont{5}{6.0}{\rmdefault}{\mddefault}{\updefault}{\color[rgb]{0,0,0}LUT $\phi_1$}%
      }}}}
      \put(6751,-3346){\makebox(0,0)[b]{\smash{{\SetFigFont{5}{6.0}{\rmdefault}{\mddefault}{\updefault}{\color[rgb]{0,0,0}LUT $\phi_2$}%
      }}}}
      \put(4456,-2356){\makebox(0,0)[lb]{\smash{{\SetFigFont{6}{7.2}{\rmdefault}{\mddefault}{\updefault}{\color[rgb]{0,0,0}$x_2$}%
      }}}}
      \put(4456,-2176){\makebox(0,0)[lb]{\smash{{\SetFigFont{6}{7.2}{\rmdefault}{\mddefault}{\updefault}{\color[rgb]{0,0,0}$x_1$}%
      }}}}
      \put(6997,-4291){\makebox(0,0)[b]{\smash{{\SetFigFont{5}{6.0}{\rmdefault}{\mddefault}{\updefault}{\color[rgb]{0,0,0}(a) Circuit $G_{\phi_1,\phi_2}(x_1,x_2)$ with LUTs}%
      }}}}
      \put(9877,-2420){\makebox(0,0)[lb]{\smash{{\SetFigFont{6}{7.2}{\rmdefault}{\mddefault}{\updefault}{\color[rgb]{0,0,0}$x_1$}%
      }}}}
      \put(9877,-3015){\makebox(0,0)[lb]{\smash{{\SetFigFont{6}{7.2}{\rmdefault}{\mddefault}{\updefault}{\color[rgb]{0,0,0}$x_2$}%
      }}}}
      \put(14719,-1135){\makebox(0,0)[lb]{\smash{{\SetFigFont{6}{7.2}{\rmdefault}{\mddefault}{\updefault}{\color[rgb]{0,0,0}$x_2$}%
      }}}}
      \put(14719,-955){\makebox(0,0)[lb]{\smash{{\SetFigFont{6}{7.2}{\rmdefault}{\mddefault}{\updefault}{\color[rgb]{0,0,0}$x_1$}%
      }}}}
      \put(14719,-1636){\makebox(0,0)[lb]{\smash{{\SetFigFont{6}{7.2}{\rmdefault}{\mddefault}{\updefault}{\color[rgb]{0,0,0}$y_1$}%
      }}}}
      \put(14719,-2131){\makebox(0,0)[lb]{\smash{{\SetFigFont{6}{7.2}{\rmdefault}{\mddefault}{\updefault}{\color[rgb]{0,0,0}$y_2$}%
      }}}}
      \end{picture}%
      
    \caption{Illustrating LUT synthesis application of BFS}
    \label{fig:example-LUT}
  \end{center}
\end{figure*}
The \emph{Boolean Functional Synthesis} (BFS) problem is as follows.

Given a Boolean formula $F(X,Y)$, with $X = \{x_1, \ldots, x_n\}$ and
$Y = \{y_1, \ldots, y_m\}$, synthesize Boolean formulas $\psi_1(X),
\ldots, \psi_m(X)$, such that
\begin{equation}\label{bfs_defa}
\exists Y F(X, Y) \equiv F(X, \psi_1(X), \ldots, \psi_m(X)).
\end{equation}
If we use $\Psi(X)$ to represent the vector $\langle \psi_1(X), \ldots,
\psi_m(X)\rangle$, we can write Eq.~\ref{bfs_defa} compactly as:
\begin{equation}\label{bfs_def}
\exists Y F(X, Y) \equiv F(X, \Psi(X))
\end{equation}
We refer to the $\psi_i$'s as \emph{Skolem functions}, and $\Psi$ as the
\emph{Skolem function vector}.

As an example consider the problem where $X =
\{x\}$, $Y=\{y_1, y_2\}$, and $F(x,y_1,y_2)$ is $\XOR(x, y_1,
y_2)$. Here $\XOR$ returns $\true$ iff the
number of true inputs to it is odd.
Possible solutions for $\Psi$ include
$\langle \neg x, \false \rangle$ and $\langle x, \true \rangle$.
We notice that $\psi_1(x)$ and $\psi_2(x)$ cannot be chosen independent of each
other.

\subsection{Application in Circuit Repair}

An important application of BFS is in circuit repair.
Chip designers often include a small level of programmability in
chips, to take care of subsequent changes in specification or bugs in the
original design.
This programmability is often achieved by introducing a small number
of programmable ``LUTs'' (for ``Look Up Tables'') into the chip, that
can be modified even after the chip is manufactured to realize the new
specification \cite{Jo}.
The question now is what logic do we fill in for each LUT so that the
circuit realizes the new spec.
We use a concrete example to illustrate how this can be modelled and
solved as an instance of the BFS problem.

Consider the circuit with two LUTs $\phi_1$ and $\phi_2$ shown in
Fig.~\ref{fig:example-LUT}(a).
Let us denote this circuit by $G_{\phi_1,\phi_2}(x_1,x_2)$, where
$\phi_1$ and $\phi_2$ are parameters representing the logic placed
in the two LUTs respectively.
Let us say the new spec for the circuit is $H(x_1,x_2)$ shown in
Fig.~\ref{fig:example-LUT}(b).
We would like to come up with concrete instantiations of $\phi_1$ and
$\phi_2$ such that the circuit behaves like the given spec $H$.
Let $G'(x_1,x_2,y_1,y_2)$ denote the given circuit with the LUTs
replaced by inputs $y_1,y_2$ respectively.
We can now consider the BFS instance $F(x_1,x_2,y_1,y_2)$ where $F$ is
the circuit $G'(x_1,x_2,y_1,y_2) \myIff H(x_1,x_2)$, as shown in
Fig.~\ref{fig:example-LUT}(c).
A solution to the given BFS problem, say $\psi_1,\psi_2$, now gives us the
required LUT instantiation.
One further step remains: we need to check that for every value of
$x_1,x_2$, there \emph{do} exist values of $y_1,y_2$ such that $F$ evaluates
to $\true$.
However this is easily done by asking a SAT solver whether the formula
$H(x_1,x_2) \myAnd \neg F(x_1,x_2,\psi_1,\psi_2)$ is unsatisfiable.

We note that in this application, the LUTs may often be constrained to
allow only a bounded number of terms (say in a POS or SOP form), or a
bounded number of inputs. Hence it is beneficial if we can synthesize
$\psi$'s with a small number of inputs and terms. Our tool BNSynth
synthesizes the compact Skolem functions $\langle \false, \neg
x_1\rangle$ for the example circuit.

\subsection{Bounded Boolean Functional Synthesis}

The application above motivates the Bounded BFS (B-BFS) problem, where,
in addition to the Boolean function $F(X,Y)$ as in Sec.~\ref{sec:bfs},
we are given a bound $K$ on the number of clauses (say in CNF or DNF
form) allowed for each Skolem function in $\Psi(X)$.
The problem now is to synthesize a solution to the BFS problem that
respects the given bound $K$.

%% The application above motivates the Bounded BFS (B-BFS) problem that 
%% asks the following question. Given a Boolean formula $F(X,Y)$ as in
%% Sec.~\ref{sec:bfs} and a bound $K$ on the maximum allowable number
%% of clauses for each Skolem function in $\Psi(X)$, synthesize Boolean
%% formulas $\psi_{1,K}(X),
%% \ldots, \psi_{m,K}(X)$, such that
%% \begin{equation}\label{bbfs_defa}
%% \exists Y F(X, Y) \equiv F(X, \psi_{1,K}(X), \ldots, \psi_{m,K}(X)).
%% \end{equation}
%% If we use $\Psi(X)$ to represent the vector $\langle \psi_1(X), \ldots,
%% \psi_m(X)\rangle$, we can write Eq.~\ref{bfs_defa} compactly as:
%% \begin{equation}\label{bbfs_def}
%% \exists Y F(X, Y) \equiv F(X, \Psi_K(X))
%% \end{equation}

%% Our tool BNSynth is, to the best of our knowledge, the first to address this problem.

%% file: sections/gcln.tex
\subsection{Gated Continuous Logic Network (G-CLN)}

\begin{figure}
    \centering
    \includegraphics{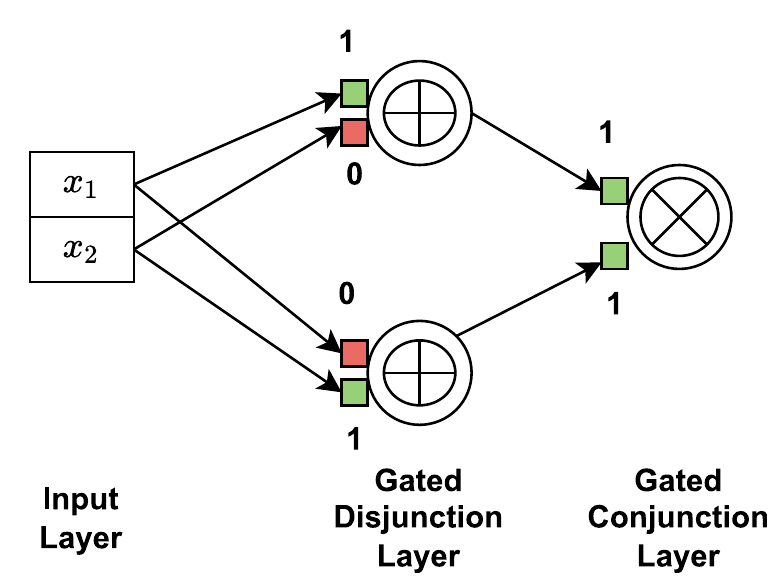}
    \caption{An instance of trained G-CLN with $K=2$ representing Boolean formula $(x_1 \wedge x_2)$, where $x_1$ and $x_2$ are Boolean variables.}
    \label{fig:gcln_ex}
\end{figure}

The enabler for BNSynth to solve the B-BFS problem is the Gated
Continuous Logic Network \cite{gcln}. This is a specialized neural
network which is used to represent logical formulae succinctly.
% \textbf{Need to say more about GCLNs}.
It uses \textit{t-norms} and \textit{t-conorms} as activation functions. \textit{T-norms} and \textit{t-conorms} \ref{gated} are continuous counterparts for \textit{logical or gate} and \textit{logical and gate}. Figure \ref{fig:gcln_ex} shows a trained instance of gcln, with CNF architecture, representing boolean formula $x_1 \wedge x_2$. 0/1 represents values of corresponding gates; 0: gate is off - don't select the input, 1: gate is on - select the input.
We refer the reader to the paper for more details; we explain the
architecture using an example in the next section.

\label{gated}
\noindent\textbf{Gated t-norms and gated t-conorms:} Given a classic t-norm $T(x, y) = x \otimes y$, we define its associated gated t-norm as 

$$T_{G}(x,y;g_{1},g_{2}) = (1*(1 - g_{1}) + x*g_{1}) \otimes (1*(1 - g_{2}) + y*g_{2})$$

Here $g_{1}, g_{2} \in [0, 1]$ are gate parameters indicating if x and y are activated, respectively.

Gates $g_{1}, g_{2}$ are learnt from Neural Network.

\noindent Given a threshold $T$ let,
\[
    g_i' =
    \begin{cases}
        1 & g_{i} > T\\
        0 & otherwise
    \end{cases}    
\]

\[\label{def}
    T_{G}(x,y;g_{1}',g_{2}')= 
\begin{cases}
    x \otimes y & g_{1}' = 1 \hspace{2mm} \text{and} \hspace{2mm} g_{2}' = 1\\
    x & g_{1}' = 1 \hspace{2mm} \text{and} \hspace{2mm} g_{2}' = 0\\
    y & g_{1}' = 0 \hspace{2mm} \text{and} \hspace{2mm} g_{2}' = 1\\
    1 & g_{1}' = 0 \hspace{2mm} \text{and} \hspace{2mm} g_{2}' = 0
\end{cases}
\]

\noindent $(1 + g_{1}(x - 1))$ gives a convex combination of $1$ and $x$ for the values of $g_{1}$ and $g_{2}$ $\in (0, 1)$.
Similarly, for $(1 + g_{1}(y - 1))$

Using DeMorgan’s laws $x \otimes y = 1 - ((1 - x) \otimes (1 - y))$, we define gated t-conorms as

$$T_{G}'(x,y;g_{1},g_{2}) = 1 - ((1 - g_{1}*x) \otimes (1 - g_{2}*y))$$,

and has following property - 
\[
    T_{G}'(x,y;g_{1}',g_{2}')= 
\begin{cases}
    x \otimes y & g_{1}' = 1 \hspace{2mm} \text{and} \hspace{2mm} g_{2}' = 1\\
    x & g_{1}' = 1 \hspace{2mm} \text{and} \hspace{2mm} g_{2}' = 0\\
    y & g_{1}' = 0 \hspace{2mm} \text{and} \hspace{2mm} g_{2}' = 1\\
    0 & g_{1}' = 0 \hspace{2mm} \text{and} \hspace{2mm} g_{2}' = 0
\end{cases}
\]

% Although POS form is rarely used in circuit synthesis (due to the asymmetric characteristics of NMOS and PMOS), it is widely used in SAT solving and Boolean reasoning.

%% file: sections/tool-architecture.tex
\section{Tool Architecture}

% \textbf{DD: Need to give the overall algo we follow. Can be taken from
%   Ravi's thesis.}

Fig.~\ref{fig:bnsynth-architecture} describes the overall architecture of BNSynth. BNSynth allows users to specify the bound on the maximum number of clauses in a Skolem function in the BFS problem. We present our overall algorithm in Algorithm \ref{alg:procedure} (B-BFS). The algorithm takes upper bound on the number of clauses $\mathit{K}$ and finds a formula ($\Psi$) within that bound satisfying the training samples in table $\mathit{T}$. Formula is obtained from the trained model using a formula extraction algorithm called \textit{FExtract}. We check the validity of $\Psi$ by constructing an error formula $\mathit{E(X, Y, Y')}$ and checking it for satisfiability. If $\Psi$ is a valid skolem function i.e. the error formula is $\mathit{UNSAT}$ then the algorithm stops else it starts the counter-example guided training loop.  To the best of our knowledge, no other tool in this space offers this flexibility. We describe the various components of the tool below.

\begin{algorithm} \caption{B-BFS}
\label{alg:procedure}
\begin{algorithmic}[1]
\State \textbf{Input:} BFS problem $F(X, Y)$.
\State \textbf{Output:} Solution to BFS problem.
\State Generate partial entries in table $T$.
\State $K$ is the number of clauses allowed by model.
\While{true}
\State Learn $\Psi$ consistent with $T$, using GCLN($K$) model. 
\State Extract the Skolem Function Vector $\Psi$ from the trained model using \textit{FExtract}.
\If{error formula $E(X, Y, Y')$ corresponding to $\Psi$ is $\mathit{UNSAT}$}
\State \Return $\Psi$ as solution.
\Else{}
\State Add counter-example given by solver to $T$.
\EndIf
\EndWhile
\State \Return
\end{algorithmic}
\end{algorithm}

% BNSynth uses a neural counterexample guided framework to synthesize sound Skolem functions.

\input{images/tool-architecture.tikz}

\subsection{Input}

BNSynth takes as input a BFS specification and a bound on the maximum clause size per Skolem function.

\noindent
\textbf{BFS Specification:} This is a Boolean relation $F(X,Y)$ over the set of input variables $X$ and the set of output variables $Y$. For example, if $X = \{ x_0\}$ and $Y = \{ y_0\}$, a possible specification is $F(X,Y) = x_0 \ \vee \ y_0$.

\noindent
\textbf{Bound on Clauses (K): } We can specify the maximum bound on the number of clauses to be synthesized for all Skolem functions in a given problem. In this paper, we consider Boolean formulas in Conjunctive Normal Form (CNF). Such formulas are conjunctions of clauses where each clause is a disjunction of literals. A literal can be an input variable $\in X$ or its negation.  

If $K=1$, a possible Skolem function $\Psi(X)$ that solves the BFS problem for the above specification is $\Psi(X) = y_0 = True$.

% BNSynth also supports synthesizing Skolem functions directly in Disjunctive Normal Form (DNF). However, we refer the readers to our tool documentation on GitHub for the same.

\subsection{Sampling} We use the weighted sampler from the tool Manthan \cite{Manthan} to sample positive points from the specification $F(X,Y)$ (i.e., points that satisfy $F$). As an extra step, we refine the samples generated by this sampler to remove non-deterministic rows. For example, in the specification, $F(X,Y) = x_0 \ \vee \ y_0$, if the sampler samples the points $(1,0)$ and $(1,1)$, these points may impede learning as it can non-deterministically assign either $0$ or $1$ to the input $1$. Thus, we randomly eliminate one of the points. Similarly, we also eliminate don't care points from the sampled data points (i.e., input points where all output values satisfy the specification), as they are not necessary conditions for our synthesis procedure.

% \textcolor{red}{Add more details on the weighted sampler from Manthan. Add the comparison as to why it is better}

\subsection{Training} This phase consists of training and formula extraction. \subsubsection{Training:} 
\begin{figure}[ht!]
	\centering
    \includegraphics[scale=0.8]{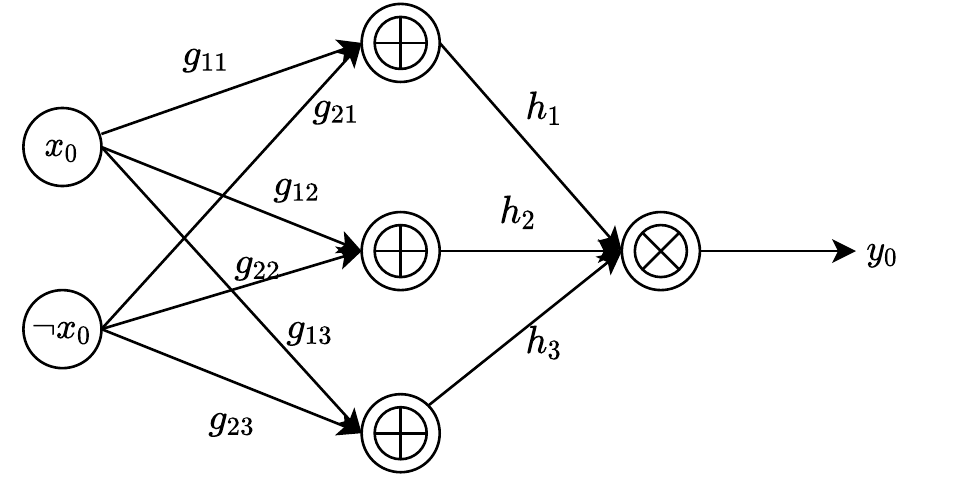}
    \caption{G-CLN CNF Architecture for $K$ = $3$ and $X$ = $\{x_0\}$}
    \label{fig:arch-1}
\end{figure}

BNSynth views the synthesis problem as a regression problem and uses a Gated Continuous Logic Network (G-CLN) \cite{gcln} as the underlying model. For our example with $K = 3$, the G-CLN architecture is shown in Figure \ref{fig:arch-1}. The weights on the edges represent the gates that will range over $\{0,1\}$ after training is complete. The activation functions in the middle layer represent the logical disjunction operator in the continuous domain and the activation function in the last layer represent logical conjunction in the continuous domain. The input layer is of size $2|X|$ as it includes negative literals as well. We train the model up to $100\%$ accuracy over the samples and then read the weights for formula extraction. For multiple outputs, this process occurs in sequence. We have refined architectures that avoid this redundancy and we describe them in our tool documentation.
\subsubsection{Formula Extraction:} After training for the given example, BNSynth learns $g_{12} = g_{22} = h_{2} = 1$, and the rest of the weights as $0$. This results in the formula $x_0 \vee \neg x_0$ which represents $True$ after simplification. This is assigned to output variable $y_0$.

% There are many representations for Boolean formulae with many applications. POS representation is widely used in SAT solving and Boolean reasoning whereas SOP is useful in two level logic optimization. BNSynth can synthesize bounded formulae in both forms. However, for brevity, we consider the POS or CNF form in this paper.

\subsection{Verification}

This phase verifies the candidate Skolem functions generated by the training phase and is in the same spirit as Manthan's verifier \cite{Manthan}.
Consider the formula $E(X,Y,Y')$, given by:
\begin{equation}
  \label{error}
    F(X, Y) \myAnd \neg F(X, Y') \myAnd (Y' \myIff \Psi).
\end{equation}

Once we generate a candidate Skolem functions $\Psi(X)$ after training, the verifier plugs it into Equation \ref{error} to check for satisfiablility. If UNSAT, the formula is valid. Otherwise, the formula is invalid. In the latter case, a counterexample is generated and added to the current sample set and training is restarted. This counterexample loop continues until valid Skolem functions are found. In our example, $y_0 = \true$ is a valid Skolem function that satisfies $F(X,Y)$.

% \textcolor{red}{Add more details on the verifier from Manthan. Add the comparison as to why it is better}

In principle, our approach terminates due to the heuristics that we use.

%% file: images/tool-architecture.tikz
\begin{figure}[htp!]
\centering

\tikzset{every picture/.style={line width=0.75pt}} %set default line width to 0.75pt        

\begin{tikzpicture}[x=0.75pt,y=0.75pt,yscale=-0.7,xscale=0.7]
%uncomment if require: \path (0,372); %set diagram left start at 0, and has height of 372

%Rounded Rect [id:dp09072259691333984] 
\draw   (244,170.27) .. controls (244,165.52) and (247.85,161.67) .. (252.6,161.67) -- (423.07,161.67) .. controls (427.82,161.67) and (431.67,165.52) .. (431.67,170.27) -- (431.67,196.07) .. controls (431.67,200.82) and (427.82,204.67) .. (423.07,204.67) -- (252.6,204.67) .. controls (247.85,204.67) and (244,200.82) .. (244,196.07) -- cycle ;
%Rounded Rect [id:dp11809394488342062] 
\draw   (342.67,246.27) .. controls (342.67,241.52) and (346.52,237.67) .. (351.27,237.67) -- (481.07,237.67) .. controls (485.82,237.67) and (489.67,241.52) .. (489.67,246.27) -- (489.67,272.07) .. controls (489.67,276.82) and (485.82,280.67) .. (481.07,280.67) -- (351.27,280.67) .. controls (346.52,280.67) and (342.67,276.82) .. (342.67,272.07) -- cycle ;
%Rounded Rect [id:dp24197934817384792] 
\draw   (243,96.27) .. controls (243,91.52) and (246.85,87.67) .. (251.6,87.67) -- (422.07,87.67) .. controls (426.82,87.67) and (430.67,91.52) .. (430.67,96.27) -- (430.67,122.07) .. controls (430.67,126.82) and (426.82,130.67) .. (422.07,130.67) -- (251.6,130.67) .. controls (246.85,130.67) and (243,126.82) .. (243,122.07) -- cycle ;
%Straight Lines [id:da48615734267129795] 
\draw    (334,132) -- (334,157) ;
\draw [shift={(334,160)}, rotate = 270] [fill={rgb, 255:red, 0; green, 0; blue, 0 }  ][line width=0.08]  [draw opacity=0] (7.14,-3.43) -- (0,0) -- (7.14,3.43) -- cycle    ;
%Straight Lines [id:da6661760709470632] 
\draw    (415,205) -- (415,234) ;
\draw [shift={(415,237)}, rotate = 270] [fill={rgb, 255:red, 0; green, 0; blue, 0 }  ][line width=0.08]  [draw opacity=0] (7.14,-3.43) -- (0,0) -- (7.14,3.43) -- cycle    ;
%Straight Lines [id:da6831663851318346] 
\draw    (418,281) -- (418,316.33) ;
\draw [shift={(418,319.33)}, rotate = 270] [fill={rgb, 255:red, 0; green, 0; blue, 0 }  ][line width=0.08]  [draw opacity=0] (7.14,-3.43) -- (0,0) -- (7.14,3.43) -- cycle    ;
%Rounded Rect [id:dp7367574482002719] 
\draw   (216,23.27) .. controls (216,18.52) and (219.85,14.67) .. (224.6,14.67) -- (359.4,14.67) .. controls (364.15,14.67) and (368,18.52) .. (368,23.27) -- (368,49.07) .. controls (368,53.82) and (364.15,57.67) .. (359.4,57.67) -- (224.6,57.67) .. controls (219.85,57.67) and (216,53.82) .. (216,49.07) -- cycle ;
%Straight Lines [id:da17650333014995345] 
\draw    (285,57.5) -- (285,83.5) ;
\draw [shift={(285,86.5)}, rotate = 270] [fill={rgb, 255:red, 0; green, 0; blue, 0 }  ][line width=0.08]  [draw opacity=0] (7.14,-3.43) -- (0,0) -- (7.14,3.43) -- cycle    ;
%Shape: Rectangle [id:dp5130564062357217] 
\draw  [color={rgb, 255:red, 155; green, 155; blue, 155 }  ,draw opacity=1 ][dash pattern={on 4.5pt off 4.5pt}] (200.67,71.33) -- (559,71.33) -- (559,298.33) -- (200.67,298.33) -- cycle ;
%Curve Lines [id:da32071172121964175] 
\draw    (341.67,259.33) .. controls (309.16,262.29) and (282.48,247.78) .. (287.42,209.12) ;
\draw [shift={(287.67,207.33)}, rotate = 98.53] [color={rgb, 255:red, 0; green, 0; blue, 0 }  ][line width=0.75]    (10.93,-3.29) .. controls (6.95,-1.4) and (3.31,-0.3) .. (0,0) .. controls (3.31,0.3) and (6.95,1.4) .. (10.93,3.29)   ;
%Rounded Rect [id:dp6858536608711236] 
\draw   (392,22.27) .. controls (392,17.52) and (395.85,13.67) .. (400.6,13.67) -- (530.4,13.67) .. controls (535.15,13.67) and (539,17.52) .. (539,22.27) -- (539,48.07) .. controls (539,52.82) and (535.15,56.67) .. (530.4,56.67) -- (400.6,56.67) .. controls (395.85,56.67) and (392,52.82) .. (392,48.07) -- cycle ;
%Curve Lines [id:da4406339811494413] 
\draw    (458,56.5) .. controls (458,106.94) and (462.7,169.61) .. (436.52,183.37) ;
\draw [shift={(434,184.5)}, rotate = 339.23] [fill={rgb, 255:red, 0; green, 0; blue, 0 }  ][line width=0.08]  [draw opacity=0] (8.93,-4.29) -- (0,0) -- (8.93,4.29) -- cycle    ;

% Text Node
\draw (332.82,182.5) node   [align=left] {Training};
% Text Node
\draw (415.69,259.77) node   [align=left] {Verification};
% Text Node
\draw (334.27,108.5) node   [align=left] {Sampling};
% Text Node
\draw (232,27) node [anchor=north west][inner sep=0.75pt]   [align=left] {BFS Specification};
% Text Node
\draw (342,139) node [anchor=north west][inner sep=0.75pt]  [font=\footnotesize] [align=left] {\textcolor[rgb]{0.5,0.5,0.5}{Samples}};
% Text Node
\draw (381,325) node [anchor=north west][inner sep=0.75pt]  [font=\footnotesize] [align=left] {\textbf{Valid} + \\\textcolor[rgb]{0.5,0.5,0.5}{Skolem Function}};
% Text Node
\draw (440,198) node [anchor=north west][inner sep=0.75pt]  [font=\footnotesize] [align=left] {\textcolor[rgb]{0.5,0.5,0.5}{Candidate Skolem }\\\textcolor[rgb]{0.5,0.5,0.5}{Function}};
% Text Node
\draw (472,83) node [anchor=north west][inner sep=0.75pt]  [font=\footnotesize,color={rgb, 255:red, 155; green, 155; blue, 155 }  ,opacity=1 ] [align=left] {\textbf{{\Large BNSynth}}};
% Text Node
\draw (237,249) node [anchor=north west][inner sep=0.75pt]  [font=\footnotesize] [align=left] {\textbf{Invalid} + \\\textcolor[rgb]{0.5,0.5,0.5}{Counterexample}};
% Text Node
\draw (404,27) node [anchor=north west][inner sep=0.75pt]   [align=left] {Bound on Clauses};

\end{tikzpicture}

\caption{BNSynth Tool Architecture}
\label{fig:bnsynth-architecture}
\end{figure}

%% file: sections/evaluation.tex
\section{Evaluation}

To evaluate BNSynth, we ask the following research questions:

\begin{itemize}
    \item \textbf{Convergence:} Does BNSynth scale as the number of variables increase?
    \item \textbf{Conciseness:} Does BNSynth synthesize smaller sized Skolem functions as compared to state of the art tools?
\end{itemize}

\noindent\textbf{Benchmarks:}
We evaluate BNSynth on custom benchmarks shown in Table \ref{tab:custom}. The custom benchmarks are constraints with smaller number of inputs and outputs, of the order of 1-6 variables and 1-26 variables respectively. They are manually created to check the soundness and conciseness of our approach. We also consider variations of LUT problems from the EPFL benchmark suite \cite{amaru2015epfl} and the ISCAS85 \cite{isca85} benchmark suite as seen in Table \ref{tab:lut}.

% The standard benchmarks are used to test the convergence of our approach. These are large  benchmarks where the input size range from 8 to 65037 variables and the output size ranges from 8 to 50039 variables. These standard benchmarks are taken from \cite{Manthan} which is the union of all the benchmarks employed in the most recent works \cite{4, akshay},which includes 609 benchmarks from different sources: Prenex-2QBF track of QBFEval-17 \cite{prenex}, QBFEval-18 \cite{prenex2}, disjunctive \cite{tacas},
% arithmetic \cite{factored} and factorization \cite{tacas}.

\noindent\textbf{Evaluation Environment:} We perform our experiments on a machine with an Intel i5-6400 processor, 8GB RAM and no GPU support.

\noindent\textbf{System Configuration:} We implemented our method with the PyTorch Framework and use the PicoSAT solver to validate the correctness of the synthesized Skolem functions. We use the Adam optimizer with learning rate 0.01 and decay set to 0. We do not set a maximum epoch as we aim to reach a $100\%$ accuracy on all benchmarks. We use the L1 regularization set to $10^{-6}$ on the G-CLN gates.

% however, we only required this for speeding up convergence in a specific sub-class of the standard benchmarks which were tough.

\noindent\textbf{Experimental Setup:}
In our preliminary experiments, we compare against the state of the art tool Manthan \cite{Manthan} in terms of time (\textbf{T}), total number of clauses (\textbf{C}), literals (\textbf{L}) and unique inputs (\textbf{I}) in the synthesized Skolem function vector.  We perform the experiments as follows. We first run BNSynth and Manthan on a specific benchmark to get $\Psi(X)$. BNSynth can give the outputs directly in either CNF or DNF, in Z3py format. In this paper, we discuss the results for CNF. However, we observed that Manthan does not synthesize the skolem functions in any normal form. Thus, we first convert the skolem functions given by Manthan in Verilog, to Z3Py, using our Verilog to Z3Py translator \footnote{\url{https://github.com/rrlcs/verilog-to-z3py}}. We then convert these Z3 formulae returned by both Manthan and BNSynth into CNF using the Z3 tactic \textit{tseitin- cnf}. Furthermore, we use an inbuilt Z3 tactic \textit{ctx-solver-simplify}, which is a powerful context based formula simplifier, repeatedly to further simplify the CNF forms returned by both tools. We have observed that this simplification maintains the CNF form and also gives the minimal formula as the simplification tactic is repeatedly applied until no more changes are possible. We have observed that the CNF formula that BNSynth generates does not minimize significantly even after using the above conversions.  For experimentation, we change the value of \textbf{K} and timeout dynamically. We iterate over \textbf{K} and timeout in the range [(1, 60), (5, 120) , (20, 120) , (50, 180) , (500, 300), (1000, 600)], where the timeout is in seconds. 

\noindent\textbf{Results:}
Table \ref{tab:custom} shows the results on the custom benchmarks. These benchmarks are named in the format $N\_I\_O$ where $N$ is the name of the benchmark, $I$ is the number of inputs and $O$ is the number of outputs. In this table, $K$ denotes the minimum value in the range [1,5,20,50,500, 1000] for which we could solve the BFS problem. We mark the rows in green for which we have $>2X$ improvement in $C$ and $L$ over Manthan. We mark in red the rows where we do not perform better than Manthan for both $C$ and $L$. It is worth noting that for all of these benchmarks, we perform at least as well as Manthan on $I$; in cases possible, we even use lesser number of inputs. However, this comes at a cost of time, as seen in the $T$ columns. On an average we see \textbf{24.2X improvement} in $C$ and \textbf{35.1X improvement} in $L$, over Manthan, for these custom benchmarks. Table \ref{tab:lut} represent real world LUT benchmarks used in the logic synthesis domain. We observe \textbf{3.2X improvement} in $C$ and \textbf{4.2X improvement} in $L$, over Manthan, for these benchmarks.

% This empirically proves the potential of a bounded BFS approach over a regular BFS approach.

\input{tables/table_custom}
\input{tables/table_lut}

% \noindent
% \textbf{Benchmarks:} As discussed, we have two benchmarks. We will refer to the first set of benchmarks as \textbf{custom benchmarks}. There are a total of 25 custom benchmarks. Table \ref{tab:custom} gives the details of custom benchmarks used for experiments. The second set is the standard benchmarks. It has a total of 609 benchmarks. We will refer to these second set of benchmarks as \textbf{standard benchmarks}. All the benchmarks are Boolean circuits written in verilog/DIMACS format.  

% \noindent \textbf{Architectures: } 
% In Section \ref{archs}, we introduced three architectures of varying complexities. We currently have three neural architectures for experimental evaluation. We focus on the first architecture. The experimentation for remaining architectures are similar for benchmarks of the size that we are considering; more details can be found in our extended version.

% Note: We count number of clauses where True False and literal is considered to be a single clause.

%% file: tables/table_custom.tex
\begin{table}[!htp]\centering
\caption{Evaluation of BNSynth on Custom Benchmarks}\label{tab:custom}
\scriptsize
\begin{tabular}{lrrrrrrrrrr}\toprule
\multirow{2}{*}{\textbf{Benchmark}} &\multicolumn{5}{c}{\textbf{BNSynth}} &\multicolumn{4}{c}{\textbf{Manthan}} \\\cmidrule{2-10}
&\textbf{K} &\textbf{T} &\textbf{C} &\textbf{L} &\textbf{I} &\textbf{T} &\textbf{C} &\textbf{L} &\textbf{I} \\\midrule
\cellcolor[HTML]{c6efce}\textbf{adder\_4\_9.v} &\cellcolor[HTML]{c6efce}20 &\cellcolor[HTML]{c6efce}58 &\cellcolor[HTML]{c6efce}27 &\cellcolor[HTML]{c6efce}68 &\cellcolor[HTML]{c6efce}4 &\cellcolor[HTML]{c6efce}2 &\cellcolor[HTML]{c6efce}1176 &\cellcolor[HTML]{c6efce}5093 &\cellcolor[HTML]{c6efce}4 \\
\textbf{mirror\_20\_20.v} &5 &88 &20 &20 &20 &2 &20 &20 &20 \\
\cellcolor[HTML]{ffc7ce}\textbf{multiplexer\_3\_3.v} &\cellcolor[HTML]{ffc7ce}5 &\cellcolor[HTML]{ffc7ce}3 &\cellcolor[HTML]{ffc7ce}8 &\cellcolor[HTML]{ffc7ce}14 &\cellcolor[HTML]{ffc7ce}3 &\cellcolor[HTML]{ffc7ce}2 &\cellcolor[HTML]{ffc7ce}6 &\cellcolor[HTML]{ffc7ce}7 &\cellcolor[HTML]{ffc7ce}3 \\
\cellcolor[HTML]{c6efce}\textbf{misc1\_2\_1.v} &\cellcolor[HTML]{c6efce}1 &\cellcolor[HTML]{c6efce}1 &\cellcolor[HTML]{c6efce}1 &\cellcolor[HTML]{c6efce}1 &\cellcolor[HTML]{c6efce}1 &\cellcolor[HTML]{c6efce}2 &\cellcolor[HTML]{c6efce}2 &\cellcolor[HTML]{c6efce}2 &\cellcolor[HTML]{c6efce}2 \\
\cellcolor[HTML]{c6efce}\textbf{misc2\_3\_1.v} &\cellcolor[HTML]{c6efce}1 &\cellcolor[HTML]{c6efce}1 &\cellcolor[HTML]{c6efce}1 &\cellcolor[HTML]{c6efce}1 &\cellcolor[HTML]{c6efce}1 &\cellcolor[HTML]{c6efce}2 &\cellcolor[HTML]{c6efce}2 &\cellcolor[HTML]{c6efce}4 &\cellcolor[HTML]{c6efce}3 \\
\cellcolor[HTML]{c6efce}\textbf{misc3\_2\_3.v} &\cellcolor[HTML]{c6efce}5 &\cellcolor[HTML]{c6efce}16 &\cellcolor[HTML]{c6efce}3 &\cellcolor[HTML]{c6efce}3 &\cellcolor[HTML]{c6efce}1 &\cellcolor[HTML]{c6efce}2 &\cellcolor[HTML]{c6efce}13 &\cellcolor[HTML]{c6efce}26 &\cellcolor[HTML]{c6efce}2 \\
\cellcolor[HTML]{c6efce}\textbf{xnor\_6\_2.v} &\cellcolor[HTML]{c6efce}50 &\cellcolor[HTML]{c6efce}43 &\cellcolor[HTML]{c6efce}51 &\cellcolor[HTML]{c6efce}232 &\cellcolor[HTML]{c6efce}6 &\cellcolor[HTML]{c6efce}3 &\cellcolor[HTML]{c6efce}160 &\cellcolor[HTML]{c6efce}594 &\cellcolor[HTML]{c6efce}6 \\
\cellcolor[HTML]{c6efce}\textbf{xor\_2\_4.v} &\cellcolor[HTML]{c6efce}5 &\cellcolor[HTML]{c6efce}49 &\cellcolor[HTML]{c6efce}5 &\cellcolor[HTML]{c6efce}7 &\cellcolor[HTML]{c6efce}2 &\cellcolor[HTML]{c6efce}2 &\cellcolor[HTML]{c6efce}15 &\cellcolor[HTML]{c6efce}28 &\cellcolor[HTML]{c6efce}2 \\
\cellcolor[HTML]{c6efce}\textbf{xor\_3\_3.v} &\cellcolor[HTML]{c6efce}20 &\cellcolor[HTML]{c6efce}14 &\cellcolor[HTML]{c6efce}6 &\cellcolor[HTML]{c6efce}14 &\cellcolor[HTML]{c6efce}3 &\cellcolor[HTML]{c6efce}2 &\cellcolor[HTML]{c6efce}91 &\cellcolor[HTML]{c6efce}309 &\cellcolor[HTML]{c6efce}3 \\
\cellcolor[HTML]{c6efce}\textbf{xor\_4\_2.v} &\cellcolor[HTML]{c6efce}20 &\cellcolor[HTML]{c6efce}15 &\cellcolor[HTML]{c6efce}9 &\cellcolor[HTML]{c6efce}33 &\cellcolor[HTML]{c6efce}4 &\cellcolor[HTML]{c6efce}2 &\cellcolor[HTML]{c6efce}57 &\cellcolor[HTML]{c6efce}213 &\cellcolor[HTML]{c6efce}4 \\
\cellcolor[HTML]{c6efce}\textbf{xor\_4\_3.v} &\cellcolor[HTML]{c6efce}20 &\cellcolor[HTML]{c6efce}14 &\cellcolor[HTML]{c6efce}17 &\cellcolor[HTML]{c6efce}45 &\cellcolor[HTML]{c6efce}4 &\cellcolor[HTML]{c6efce}2 &\cellcolor[HTML]{c6efce}119 &\cellcolor[HTML]{c6efce}523 &\cellcolor[HTML]{c6efce}4 \\
\cellcolor[HTML]{c6efce}\textbf{xor\_5\_2.v} &\cellcolor[HTML]{c6efce}20 &\cellcolor[HTML]{c6efce}12 &\cellcolor[HTML]{c6efce}17 &\cellcolor[HTML]{c6efce}81 &\cellcolor[HTML]{c6efce}5 &\cellcolor[HTML]{c6efce}3 &\cellcolor[HTML]{c6efce}141 &\cellcolor[HTML]{c6efce}470 &\cellcolor[HTML]{c6efce}5 \\
\cellcolor[HTML]{c6efce}\textbf{xor\_5\_3.v} &\cellcolor[HTML]{c6efce}20 &\cellcolor[HTML]{c6efce}55 &\cellcolor[HTML]{c6efce}21 &\cellcolor[HTML]{c6efce}82 &\cellcolor[HTML]{c6efce}5 &\cellcolor[HTML]{c6efce}2 &\cellcolor[HTML]{c6efce}546 &\cellcolor[HTML]{c6efce}2628 &\cellcolor[HTML]{c6efce}5 \\
\cellcolor[HTML]{c6efce}\textbf{xor\_6\_2.v} &\cellcolor[HTML]{c6efce}50 &\cellcolor[HTML]{c6efce}24 &\cellcolor[HTML]{c6efce}52 &\cellcolor[HTML]{c6efce}242 &\cellcolor[HTML]{c6efce}6 &\cellcolor[HTML]{c6efce}3 &\cellcolor[HTML]{c6efce}179 &\cellcolor[HTML]{c6efce}603 &\cellcolor[HTML]{c6efce}6 \\
\cellcolor[HTML]{c6efce}\textbf{xor\_6\_10.v} &\cellcolor[HTML]{c6efce}50 &\cellcolor[HTML]{c6efce}51 &\cellcolor[HTML]{c6efce}155 &\cellcolor[HTML]{c6efce}731 &\cellcolor[HTML]{c6efce}6 &\cellcolor[HTML]{c6efce}4 &\cellcolor[HTML]{c6efce}9825 &\cellcolor[HTML]{c6efce}62462 &\cellcolor[HTML]{c6efce}6 \\
\cellcolor[HTML]{c6efce}\textbf{xor\_6\_26.v} &\cellcolor[HTML]{c6efce}50 &\cellcolor[HTML]{c6efce}170 &\cellcolor[HTML]{c6efce}359 &\cellcolor[HTML]{c6efce}1888 &\cellcolor[HTML]{c6efce}6 &\cellcolor[HTML]{c6efce}5 &\cellcolor[HTML]{c6efce}94262 &\cellcolor[HTML]{c6efce}754276 &\cellcolor[HTML]{c6efce}6 \\
\textbf{xor\_8\_8.v} &1000 &404 &680 &4720 &8 &8 &1475 &6697 &8 \\
\cellcolor[HTML]{ffc7ce}\textbf{xor-implies\_8\_8.v} &\cellcolor[HTML]{ffc7ce}1000 &\cellcolor[HTML]{ffc7ce}129 &\cellcolor[HTML]{ffc7ce}568 &\cellcolor[HTML]{ffc7ce}3727 &\cellcolor[HTML]{ffc7ce}8 &\cellcolor[HTML]{ffc7ce}6 &\cellcolor[HTML]{ffc7ce}114 &\cellcolor[HTML]{ffc7ce}340 &\cellcolor[HTML]{ffc7ce}8 \\
\cellcolor[HTML]{c6efce}\textbf{xnor-implies\_6\_2.v} &\cellcolor[HTML]{c6efce}50 &\cellcolor[HTML]{c6efce}26 &\cellcolor[HTML]{c6efce}51 &\cellcolor[HTML]{c6efce}238 &\cellcolor[HTML]{c6efce}6 &\cellcolor[HTML]{c6efce}3 &\cellcolor[HTML]{c6efce}236 &\cellcolor[HTML]{c6efce}747 &\cellcolor[HTML]{c6efce}6 \\
\midrule
\multicolumn{6}{c}{\multirow{2}{*}{\textbf{Average Improvement of BNSynth over Manthan}}} &\textbf{T} &\textbf{C} &\textbf{L} &\textbf{I} \\
& & & & & &\textbf{0.3} &\textbf{24.2} &\textbf{35.1} &\textbf{1.2} \\
\bottomrule
\end{tabular}
\end{table}

%% file: tables/table_lut.tex
\begin{table}[!htp]\centering
\caption{Evaluation of BNSynth on LUT Benchmarks}\label{tab:lut}
\scriptsize
\begin{tabular}{lrrrrrrrrrr}\toprule
\multirow{2}{*}{\textbf{Benchmark}} &\multicolumn{5}{c}{\textbf{BNSynth}} &\multicolumn{4}{c}{\textbf{Manthan}} \\\cmidrule{2-10}
&\textbf{K} &\textbf{T} &\textbf{C} &\textbf{L} &\textbf{I} &\textbf{T} &\textbf{C} &\textbf{L} &\textbf{I} \\\midrule
\textbf{lut1\_2\_2.v} &5 &6 &2 &2 &1 &1 &2 &2 &1 \\
\textbf{lut2-pla\_2\_2.v} &5 &3 &2 &2 &1 &1 &2 &2 &1 \\
\textbf{lut3-c17a\_5\_2.v} &5 &1 &4 &7 &4 &1 &7 &18 &5 \\
\cellcolor[HTML]{c6efce}\textbf{lut4-c17b\_5\_2.v} &\cellcolor[HTML]{c6efce}20 &\cellcolor[HTML]{c6efce}3 &\cellcolor[HTML]{c6efce}14 &\cellcolor[HTML]{c6efce}46 &\cellcolor[HTML]{c6efce}5 &\cellcolor[HTML]{c6efce}1 &\cellcolor[HTML]{c6efce}94 &\cellcolor[HTML]{c6efce}313 &\cellcolor[HTML]{c6efce}5 \\
\cellcolor[HTML]{c6efce}\textbf{lut5-c17c\_5\_2.v} &\cellcolor[HTML]{c6efce}5 &\cellcolor[HTML]{c6efce}1 &\cellcolor[HTML]{c6efce}4 &\cellcolor[HTML]{c6efce}9 &\cellcolor[HTML]{c6efce}4 &\cellcolor[HTML]{c6efce}1 &\cellcolor[HTML]{c6efce}16 &\cellcolor[HTML]{c6efce}51 &\cellcolor[HTML]{c6efce}4 \\
\cellcolor[HTML]{c6efce}\textbf{lut6-c17d\_5\_2.v} &\cellcolor[HTML]{c6efce}5 &\cellcolor[HTML]{c6efce}1 &\cellcolor[HTML]{c6efce}4 &\cellcolor[HTML]{c6efce}8 &\cellcolor[HTML]{c6efce}4 &\cellcolor[HTML]{c6efce}1 &\cellcolor[HTML]{c6efce}15 &\cellcolor[HTML]{c6efce}48 &\cellcolor[HTML]{c6efce}4 \\
\cellcolor[HTML]{c6efce}\textbf{lut7-ctrl\_7\_1.v} &\cellcolor[HTML]{c6efce}5 &\cellcolor[HTML]{c6efce}8 &\cellcolor[HTML]{c6efce}5 &\cellcolor[HTML]{c6efce}11 &\cellcolor[HTML]{c6efce}7 &\cellcolor[HTML]{c6efce}1 &\cellcolor[HTML]{c6efce}21 &\cellcolor[HTML]{c6efce}67 &\cellcolor[HTML]{c6efce}7 \\
\midrule
\multicolumn{6}{c}{\multirow{2}{*}{\textbf{Average Improvement of BNSynth over Manthan}}} &\textbf{T} &\textbf{C} &\textbf{L} &\textbf{I} \\
& & & & & &\textbf{0.6} &\textbf{3.2} &\textbf{4.2} &\textbf{1.0} \\
\bottomrule
\end{tabular}
\end{table}